\documentclass{article}

\usepackage{arxiv}

\usepackage[utf8]{inputenc} 
\usepackage[T1]{fontenc}    
\usepackage{hyperref}       
\usepackage{url}            
\usepackage{booktabs}       
\usepackage{amsfonts}       
\usepackage{nicefrac}       
\usepackage{microtype}      
\usepackage{lipsum}
\usepackage{graphicx}
\usepackage[numbers]{natbib}
\usepackage{subcaption}

\title{Reflective Artificial Intelligence}

\author{
 Peter R. Lewis \\
  Faculty of Business and Information Technology \\
  Ontario Tech\\
  Ontario, Canada \\
  \texttt{peter.lewis@ontariotechu.ca} \\
   \And
 Ștefan Sarkadi \\
  Dept. of Informatics\\
  King's College London\\
  London, UK \\
  \texttt{stefan.sarkadi@kcl.ac.uk} \\
}

\begin{document}
\maketitle
\begin{abstract}
Artificial Intelligence (AI) is about making computers that do the sorts of things that minds can do, and as we progress towards this goal, we tend to increasingly delegate human tasks to machines. However, AI systems usually do these tasks with an unusual imbalance of insight and understanding: new, deeper insights are present, yet many important qualities that a human mind would have previously brought to the activity are utterly absent. Therefore, it is crucial to ask which features of minds have we replicated, which are missing, and if that matters. One core feature that humans bring to tasks, when dealing with the ambiguity, emergent knowledge, and social context presented by the world, is reflection. Yet this capability is utterly missing from current mainstream AI. In this paper we ask what \emph{reflective AI} might look like. Then, drawing on notions of reflection in complex systems, cognitive science, and agents, we sketch an architecture for reflective AI agents, and highlight ways forward.
\end{abstract}


\section{Introduction}

Margaret Boden has described artificial intelligence as being about making `computers that do the sorts of things that minds can do'~\citep[p1]{Boden:2016:AI}. One strength of this definition lies in the fact that it does not start from an arbitrary description of the things that might be necessary or sufficient for a system to `count' as AI, but it encourages us to ask: what are the sorts of things that our minds do? Further, a curious mind is then tempted to ask: could we replicate these things? If so, how? If not, why not and does that matter?

The definition also implies that there are things that human minds currently do, that in the future machines might do instead. This is not only true now, but as Mayor~\cite{Mayor:2018} discusses, has been the case since antiquity and likely will be far into the future. It is this transference of activity that gives rise to the seemingly constant stream of examples of new `AI technologies'. These mostly do things that human minds used to, or wished to do.
This is, of course, also the source of many of the issues and benefits that arise from the creation and use of AI technology: as we figure out how to replicate some of the things that minds can do, we delegate these things to machines. 
This typically brings increased automation, scale, and efficiency, which themselves contain the seeds of both enormous potential social and economic benefit, and potential real danger and strife.


Further, we can notice that these AI technologies usually do this with an unusual (im)balance of insight and understanding. New, deeper insight and understanding often arise from the models employed, while many of the `qualities' that a human mind would have previously brought to the activity, are utterly absent.

In designing and analysing embodied AI technologies, the concept of an \emph{intelligent agent} is central, and necessitates descriptions that are abstracted from the natural intelligences they are inspired by or modelled on. This abstraction, in turn, means that the notion of an AI agent only partially captures the mental, cognitive, and physical features of natural intelligence. Hence, it is important to ask: are the features that we have included sufficient for what is needed? Are we satisfied with leaving out those which we did?

Frank and Virginia Dignum have recently reminded us how powerful the concept of an agent is in AI \citep{dignum2020agents}. They have also pointed out some of the paradigmatic failures of \textit{agent-based modelling} (ABM) and \textit{multi-agent systems} (MAS). ABM methodologies aim to describe a large and complex system of agent populations by using analysis tools in the form of agent-based simulation. MAS methodologies, instead, focus on the operational side of interacting systems, where agents operate to create changes in their environment. However, neither methodology is fit for designing human-like agent architectures. The focus of their discussion~\citep{dignum2020agents} is to propose a social MAS architecture and argue that future socially-aware AI architectures should be different from today's common utility- and goal- driven models. A similar proposal was made by Ron Sun years ago, namely that agent methodologies need Cognitive Science and vice-versa~\citep{sun2001cognitive} to address complex socially-aware AI and be able to design such architectures. Antonio Lieto refreshes this proposal~\citep{lieto2021cognitive}. In this paper we continue this line of thought, sketching an agent architecture that captures some reflective capabilities, based on cognitive theories. Due to the complex and modular nature of reflection it is impossible to find a single unique and crisp clear definition of the term reflection~\citep{Pitt:2014:book}. Reducing the definition to a single process or component of an architecture would fail to address the richness of this cognitive process and would be counter-productive in explaining how all of the processes and components at play interact for human-like reflection to happen. Thus, in order to do it justice, similarly to \citep{tine2009uncovering}, we adopt a differentiated theory approach to convey the notion of reflection.

\section{Playing Chess Isn't Just About Chess}

So what are the sorts of things minds do? As Richard Bellman suggested in 1978 \citep{Bellman_1978}, these include  activities such as `decision-making, problem solving, learning...' And the sheer quantity of research on machines that can do these activities is astounding. Yet as Bellman's ellipsis suggests, this is clearly an incomplete list. Perhaps any such list would be.
It might be more useful to think situationally. We can ask: which features of our minds do we bring to different activities? Let us explore a thought experiment using a canonical example: chess. When playing chess, we largely bring the ability to reason, to plan ahead, to use heuristics, and to remember and recall sequences of moves, such as the caro-kann defence. Against an anonymous opponent on the Internet, we might try to use these abilities as best we can.

When playing chess with a child, however, we might typically bring a few more features too: patience, empathy (for example to understand the child's current mental model of the game to help coach them), and also some compassion, since proficient chess players could likely beat most children every time and make it less interesting all round. Letting children win is also not helpful, but a parent might play out a different sequence of moves to open up more in-game experiences from time to time. As a young player grows up, benefiting from both more brain development and experience at chess, and finds joy in different parts of the game, the way an adult opponent might do this will change. A good teacher might think back over previous games, reflect on the changes in the child's understanding and reasoning, and responses to moves. They might use this to speculate on and mentally play out possible future games.

This chess example illustrates three points: (i) even playing chess is not just about problem solving; (ii) rather unsurprisingly, our mental features are rich, contextual, and flexible; and (iii) we reflect on our situations, our current and past behaviour in them, and the likely outcome of those behaviours including the impact on others, in order to choose which mental features to engage. This is not just about flexible behaviour selection, it’s about which mechanisms – which of Boden's \emph{sorts of things} – even kick in. 
What can this teach us about how we might want to build AI systems? 
Returning to the idea that we are delegating mental activity to machines,
it tells us that perhaps we might want to have a similar ability in AI agents. 

\section{The Dangers of Incomplete Minds}

Many people tie themselves in knots trying to define `intelligence', hoping that that will lead us to somewhat of a more complete (and, they often say, more helpful) definition of `artificial intelligence'. One example of such a discussion can be found in a recent special issue of the Journal of Artificial General Intelligence~\citep{Monett_et_al:2020}. As pointed out by Sloman in that collection, much of this definitional wrangling misses the point, at least from the perspective of deciding when we want to accept a computer to replace part of the activity previously done in society by human minds. Better questions might be to ask: what can this thing \emph{do}; and what is it \emph{for}? Consider: if we are deciding to put a machine in a position where it is carrying out a task in a way that we are satisfied is equivalent to what previously only a human mind could do, we have admitted something about the nature of either the task, or the machine, or our minds.
Perhaps an AI system is simply a machine that operates sufficiently similarly to our mind, at least in some situations, that we are prepared to accept the machine operating in lieu of us. So this leads us to ask when and why we would be prepared to accept this. Or perhaps, given most AI systems (and minds) cannot be not fully understood or controlled, when and why we would be prepared to trust it to do so~\citep{Lewis_Marsh:2021}.

In one recent example, the seemingly harmless act of allowing a `smart' voice assistant to propose entertainment activities to a child led to a life-threatening suggestion from a supposed trusted AI\footnote{\url{https://www.bbc.co.uk/news/technology-59810383}}. Normally, when delegating the proposal of children's play activities, we would expect that the person we had delegated that to would have not only a decent dose of common sense, but also the ability to consider the potential consequences of any ideas that sprung to mind before vocalizing them.

In another now well-known example, Amazon's automated recruiting tool, trained on data from previous hiring decisions, discriminated based on gender for technical jobs~\citep{Amazon_AI:2018}. Here, the delegation is from professional recruiters and hiring managers to a computer that replicates (some of) the mental activity they used to do. The aims are automation, scale, and efficiency.
That such a sexist system was put into practice at all is at the very least unfortunate and negligent. It is also tempting to argue that these are `just bad apples', and that better regulation is the answer. It may be. But what is particularly interesting in our context is that people – hiring managers, shareholders, applicants – trusted the system to do something that, previously, a human mind did. But unlike the mind of the professional it replaced, it had no way of reflecting on the social or ethical consequences, or on the virtue or social value of its actions, or even if its actions were congruent with prevailing norms or values. That it had no way of reflecting on this meant that it also stood no chance at stopping or correcting itself. Indeed, neither of the above AI systems even had the mental machinery to do such a thing -- this part of the mental activity is, as yet, nowhere near delegated. This leads to an unusual divorce of accompanying mental qualities that would normally work in concert. No wonder the behaviour might seem a little pathological.
As humans, a core part of our intelligence is our ability to reflect in these ways; reflection is a core mental mechanism that we use to evaluate ourselves. The existence of this form of self-awareness and self-regulation can be key to why others may find us trustworthy. Could we expect the same of machines?

\section{The Role of Reflection in Driving Human Behaviour}
One aspect of reflection is captured by what Socrates called his \emph{daemon}~\citep{Plutarch}, something that checked him \emph{`from any act opposed to his true moral and intellectual interests'}~\citep{Plato_Shorey}. Socrates saw this as a divine signal, not proposing action, but monitoring it, and intervening if necessary. If such a check were based on morals or ethics, we might call this a conscience. If it were based on broader goals than simply the immediate (for example, choosing a chess move to make against your daughter), we might call this considering the bigger picture. Essentially, this is a process that notices what we are thinking, what we are considering doing, and allows and explores the thought, but can prevent the action. It decides whether to do this by contextualising the action. Contexts, as alluded to above, might be ethical, cultural, political, social, or based on non-immediate (higher-level, longer-term, or not immediately visible) goals.

What Socrates presents here requires a `Popperian' mind according to Dennett's Tower of Generate and Test \cite{dennett2013role}. Essentially, in what he describes as a framework for `design options for brains', Dennett notes that (at the bottom of the Tower) the testing of hypotheses is done by Darwinian evolution: hypotheses are generated through mutations and the placing of novel organisms in the world and tested through their survival. Above this, Skinnerian creatures test hypotheses by taking actions and learning in an operant conditioning fashion, based on environmental feedback within their lifetime. Higher still are Popperian and Gregorian creatures, which have the mental capability to bring hypothesis testing internally to their mind, rather than requiring it to be done in the world. Both of these operate with forms of reflection: put simply, Popperian creatures think about what to think, and Gregorian creatures, using tools, language, and culture, extend this to think about \emph{how} to think.

One plausible way these Popperian and Gregorian creatures' minds might work is Hesslow's \emph{Simulation Theory of Cognition} \citep{hesslow2002conscious,hesslow2012current}. Hesslow's hypothesis is that there exists a mechanism in the brain that helps agents reason about the consequences of their actions in an environment by simulating the stimuli of their behaviour in that environment, without having this behaviour previously reinforced by actual stimuli generated by their past behaviour. For example, this mechanism allows an agent to think about the deadly consequences of driving towards a concrete wall at high speed without having done it beforehand.    

Schön compares the inherent nature of reflection in professional practice with a purely technically rational approach that might be characterised by up-front specification and subsequent problem solving \cite{schon}. As opposed to passive problem solving, \textit{active experimentation} is emphasised by how professional practitioners deal `on-the-fly' with uncertainty, ambiguity and emergent knowledge inherent in tasks. From a technical rationality perspective, Schön argues, `professional practice is a process of problem solving', yet, `in real-world practice, problems do not present themselves to the practitioner as givens. They must be constructed from the materials of problematic situations which are puzzling, troubling, and uncertain.' This means that the sorts of things that (to continue the example) a professional recruiter does is not simply problem solving in a defined setting: there are patterns and mechanical aspects to their work, but the problem is always somewhat uncertain, and emerges from practice and the setting. Thus, we arrive at what Schön describes as `an epistemology of practice which places technical problem solving within a broader context of reflective inquiry.'

Similarly, Weinberg \cite{weinberg1972science} argues that scientific knowledge and problem solving take place within a broader, untidier, and chaotic complex world. This world contains questions that, although appearing scientific, in fact transcend science in their nature.
It is here where reflection based on experience can be a mechanism for contextualizing operational knowledge and problem solving within this trans-scientific world.

A model that captures reflection in practice, that is both exploratory and governed by a sense of the bigger picture and the principles that govern our intended direction, is Kolb's learning cycle~\citep{Kolb:1984}. His `Experiential Learning Model' comprises four phases: i) having a concrete experience, ii) an observation and subjective evaluation of that experience in context, iii) the formation of abstract concepts based upon the evaluation, and iv) the formation of an intention to test the new concepts, leading to further experience.

\begin{figure}
    \centering
    \includegraphics[scale=.4]{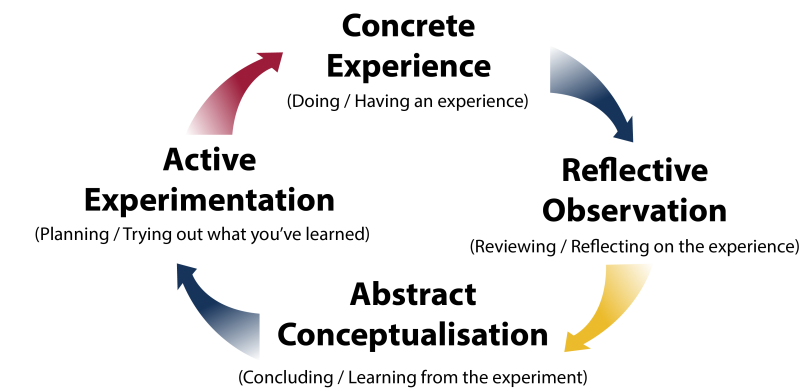}
    \caption{Kolb's `Experiential Learning Model'. Source:~\citep{Kolb:1984}. The model shows captures the cognitive cycle in humans that is responsible for learning from experience.}
    \label{fig:cycle}
\end{figure}

\section{Where are we now?}

Reflection in humans is complex and comprises numerous related phenomena. This makes it extremely difficult, if not impossible to find a single, crisp, and clear definition. This is usually the case with complex socio-cognitive phenomena (e.g., \cite{Pitt:2014:book}).
Our approach instead is to contribute to building a ‘differentiated theory', as is often done in social psychology \citep{tine2009uncovering}. This allows us to collect and compare the different ways in which phenomena all commonly referred to as being part of ‘reflection’ interact. In doing so, we aim to build towards a socio-cognitive theory of reflection in AI.

Let us first examine the current state of AI, in this light.

\begin{figure}[ht]
    \centering
    \includegraphics[width=.7\linewidth]{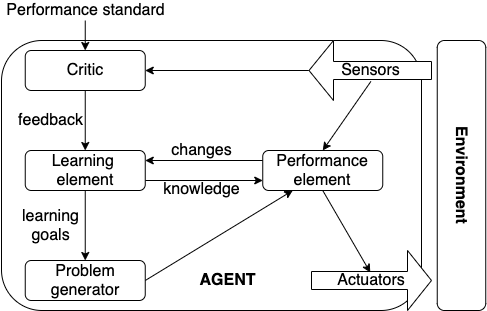}
    \caption{Critic Agent Architecture \citep{russell2021artificial}. We introduce this architecture  as a baseline AI architecture that manages to capture various aspects of perceiving, learning, planning, reasoning and acting as different qualitative processes. One can visually contrast this architecture with with the other mainstream architectures, old and new, in AI.}
    \label{fig:critic}
\end{figure}

\begin{figure*}[ht]
\centering
\begin{subfigure}[b]{0.48\textwidth}
    \includegraphics[height=6.5cm]{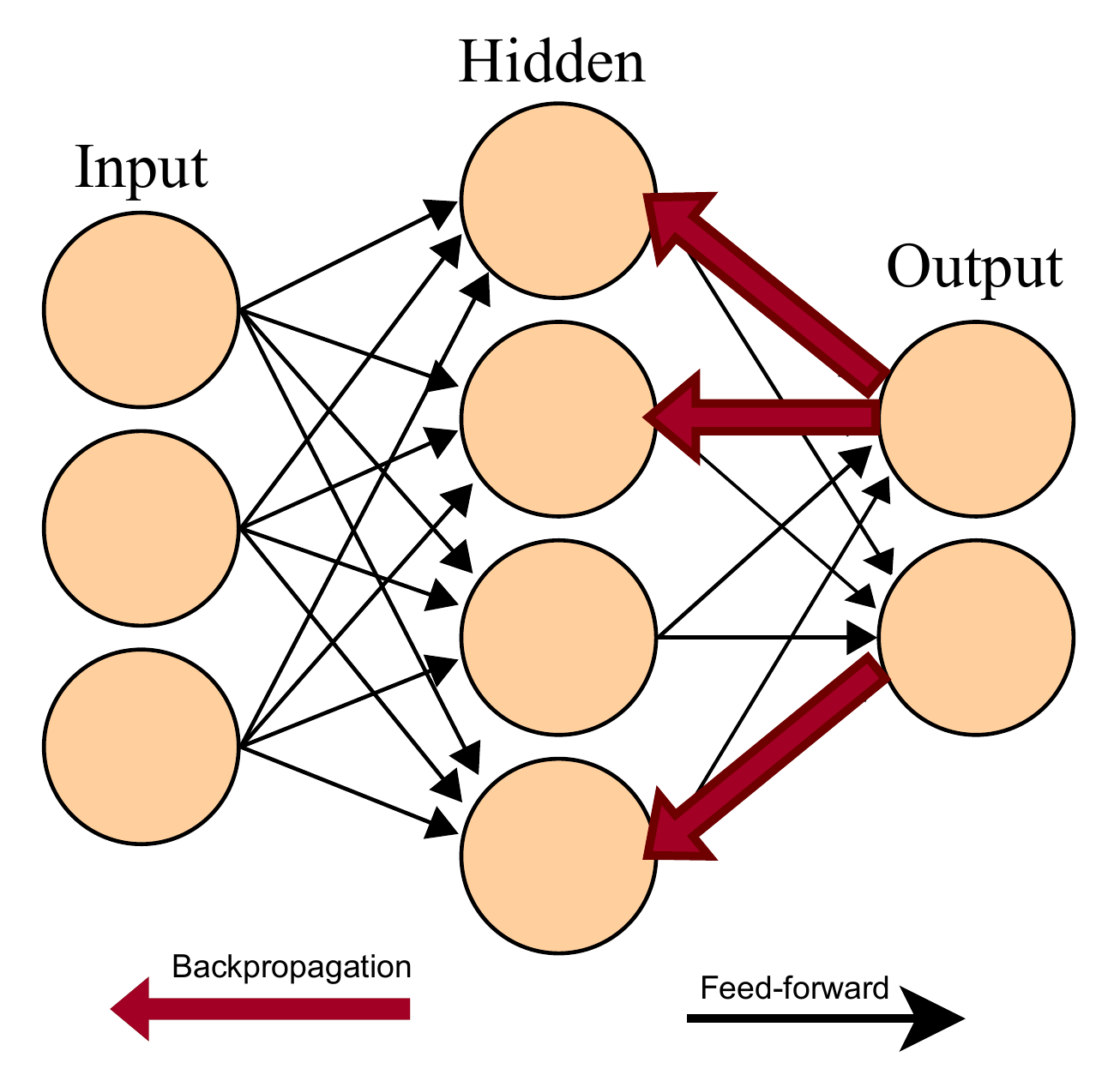}
\end{subfigure}
\begin{subfigure}[b]{0.48\textwidth}
    \includegraphics[height=6.5cm]{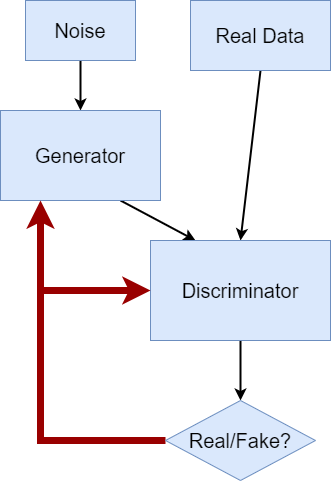}
\end{subfigure}
\caption{ANN architecture (left, by Colin Burnett from \url{https://en.wikipedia.org/wiki/Connectionism}) and GAN \citep{goodfellow2014generative}. Considering these common machine learning architectures, it is clear that there is a lack of any reflective `loop'. Although these achieve different outcomes, they are qualitatively equivalent in the sense that they both operate at a single level of abstraction when it comes to information processing. There is no self-reference: the loops in both cases are for feedback, in much the same way that the Critic Agent operates. Additionally, even though Kolb's model of experiential learning is a model of learning in humans, it also presents (albeit at a high level) qualitative processes that ANNs and GANs do not.}
\end{figure*}

\begin{figure}[ht]
    \centering
    \includegraphics[width=\linewidth]{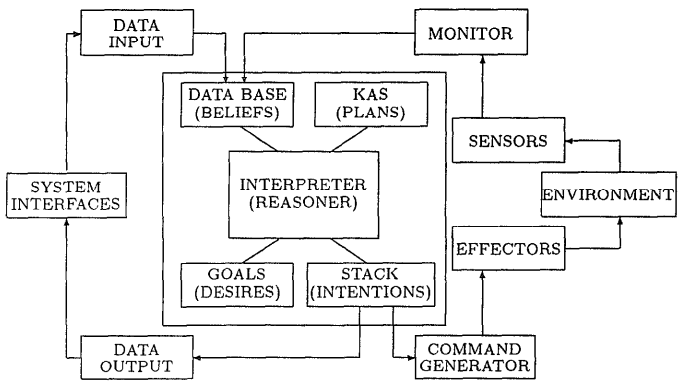}
    \caption{PRS architecture (Source: \citep{georgeff1987reactive}). The PRS architecture is similar to the Critic agent architecture in the sense that it allows us to break down different qualitative processes. The difference between the Critic architecture and the PRS is that the PRS does not include a learning component, but it has a richer representation of the processes and elements responsible with driving the reasoning behind the actions that are executed in the environment. PRS also allows for an eventual learning component to be plugged into the system interface which feeds data into the belief base.}
    \label{fig:PRS_arch}
\end{figure}

\begin{figure}[ht]
    \centering
    \includegraphics[width=.7\linewidth]{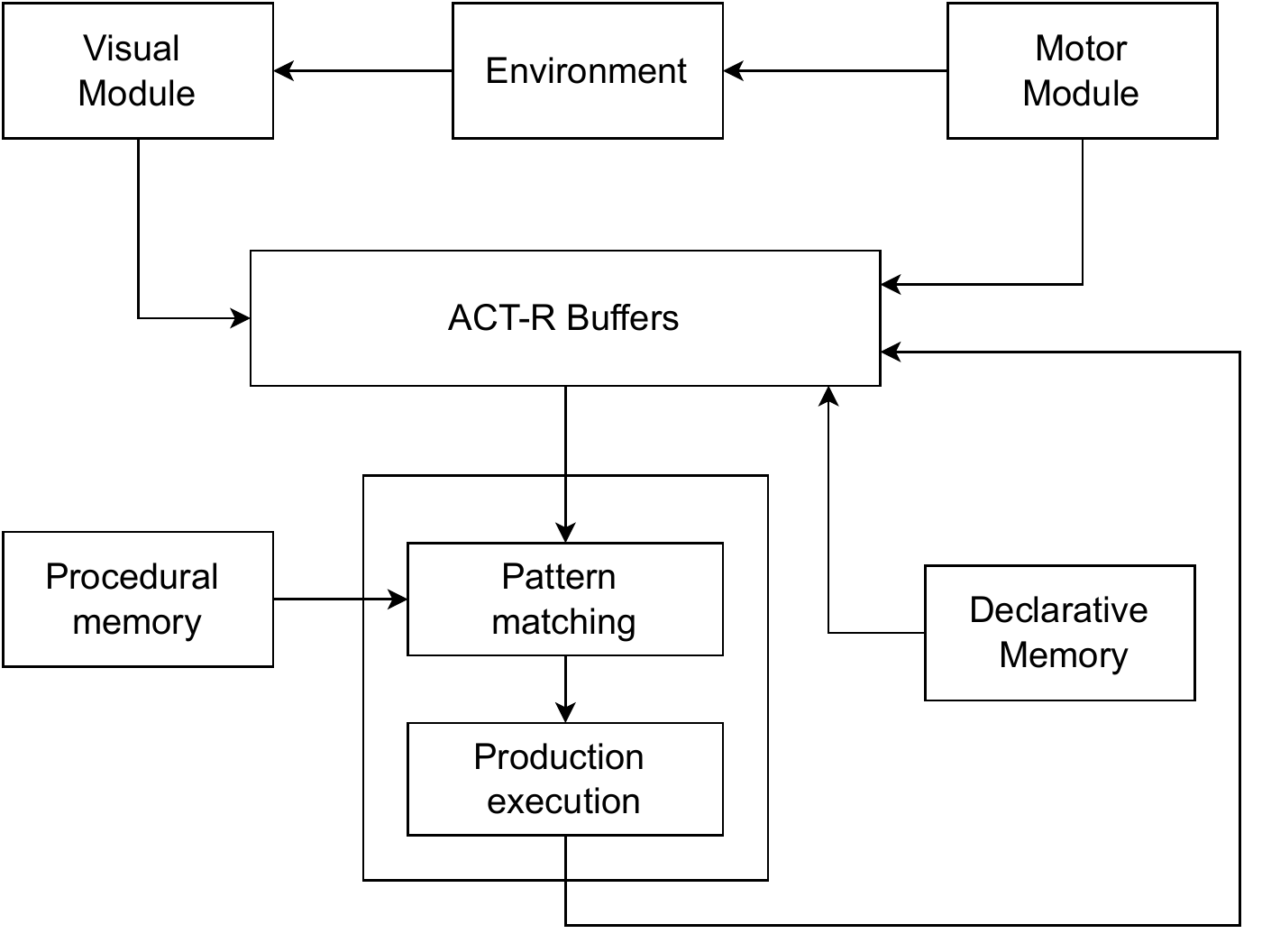}
    \caption{ACT-R architecture \citep{anderson1997act}. It is crucial to note that ACT-R is not an AI agent architecture, rather a cognitive architecture that was used as an expert system. The original purpose of ACT-R is to map and understand human cognition as a set of modular components that execute procedures to produce behaviour in a specific domain. ACT-R assumes that all cognitive components are represented and driven by declarative and procedural memory.}
    \label{fig:Act_R}
\end{figure}

\begin{figure}[ht]
    \centering
    \includegraphics[width=\linewidth]{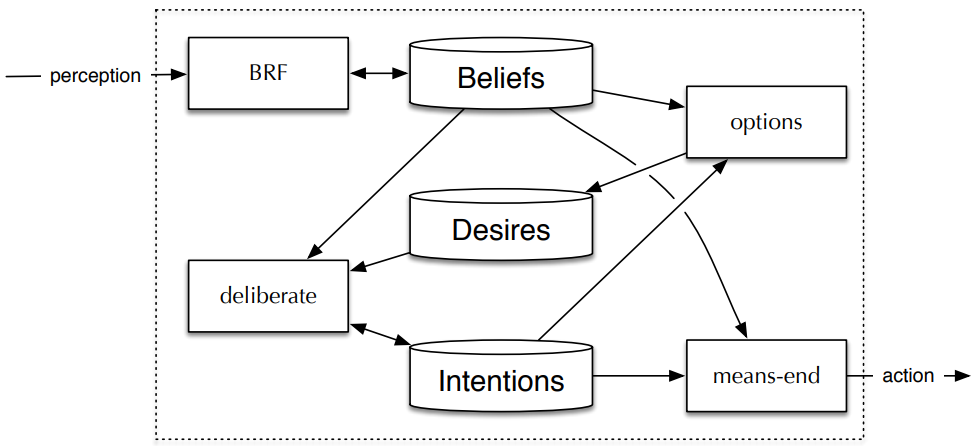}
    \caption{BDI architecture (by Jomi F. Hübner from \url{https://ai4industry.sciencesconf.org/data/Multi_Agent_Systems_lecture.pdf}). The BDI architecture was designed to help AI agent designers build intuitive and interpretable AI agents capable of practical reasoning. The architecture depicts different qualitative processes and elements responsible for meta-reasoning (deliberation) and belief revision (BRF), which then help the agent decide what to do in a given circumstance in order to achieve their goals/desires in a dynamic environment.}
    \label{fig:BDI_arch}
\end{figure}

\paragraph{Artificial Neural Network Architectures (ANN)} Initially introduced in~\cite{mcculloch1943logical} and in the form of a simple perceptron by Rosenblatt \cite{rosenblatt1958perceptron}, ANNs have gained major traction in the AI community. ANNs are highly applicable in the domain of statistical machine learning in which they are trained to perform various tasks, and outperform humans in quite a few of these tasks \citep{lecun2015deep}. However, like any supervised learning model, ANN's over-reliance on historical data means that they learn to repeat \emph{what has been done}, not \emph{what ought to be done}. Coupled with their largely black-box nature, this leads to a propagation of existing systemic biases that is difficult to identify or address. Post-hoc methods to interpret and `explain' ANN-based models, such as LIME and SHAP~\citep{samek2021explaining}, result not in an explanation of the internal mechanics of ANNs, but approximations in the form of equivalent interpretable models. For instance, in order to explain a deep-ANN, a decision tree or a heat-map are generated as an approximate function between the inputs and the outputs of the deep-ANN. This may, perhaps, be seen as a form of external, open loop reflection (but typically by others, not by the system itself); in and of themselves, the architecture of (feed-forward) ANNs has no capability for reflection.

\paragraph{Generative Adversarial Network Architectures (GAN)}

GANs~\citep{goodfellow2014generative} are one recent example of how ANNs can be used as building blocks within an explicitly designed architecture. These pitch two multi-layered perceptrons (ANNs) against each other in a 2-player minimax game. The higher-level architecture here captures the sort of competitive creative co-adaptation found within co-evolutionary systems.

When it comes to the human ability of reflection, GANs by themselves are incapable of representing the process. While their architecture contains a feedback loop, it does not operate at the meta level: the architecture is `flat'. It is not generally considered that a GAN (or coevolution in general) adds any type of high-level cognitive process. While ANNs in general are just clusters of interconnected nodes with weighted edges, and the same may be said of the brain, we contend that there are essentially two approaches to generating cognitive processes of this type: one is a complex systems approach, where the virtual machine~\citep{Sloman_Chrisley:2003,Sloman:2013,Sloman:1996:Rock} operationalizing cognition emerges through complexity; the second is through an explicit architecture, as we do in this paper.
Because AI agents that solely use ANNs such as these cannot reflect about themselves and the consequences of their actions in the world, they can behave anti-socially with no ability to know this.

One of the reflective components from Kolb's cycle (see Section \ref{sec:roleofreflection}, Fig. \ref{fig:cycle}) missing here is Active Experimentation, which is distinct from exploration. Active Experimentation is also a multifaceted process, that includes at least exploration and active learning at the meta-level, and also intentional reconceptualizations of existing knowledge, i.e. Dennett's Tower of Generate-and-Test~\citep{dennett1975law,dennett2013role}, in order to be able to reflect on the value of new models.

\paragraph{Practical Reasoning Architectures}
Procedural reflection provides a form of hard-coded first-order meta-reasoning in PRS, where a process is specified that deliberates over possible execution plans. The Procedural Reasoning System (based on Lisp \citep{smith1984reflection} - see Fig.~\ref{fig:PRS_arch}) implements this by passing symbols from the previous state (or lower symbolic level) to the current one such that we can say what the system was up to in the previous state \citep{smith1982procedural}. There is no recursive reasoning capability or ability to learn and integrate arbitrary self-models, both which are are crucial for many forms of reflection.

\paragraph{BDI Architectures}
Later, Belief-Desire-Intention (BDI) architectures \citep{rao1995bdi}, Fig.~\ref{fig:BDI_arch}, based on PRS, were introduced to structure cognitive reasoning based representations of and interactions between propositional models capturing the agent's beliefs, desires, and intentions.

Reflection has only recently been modelled in BDI architectures, again in the form of procedural reflection as in PRS.
According to a recent survey on BDI agents \citep{de2020bdi}, only one paper was identified that implemented reflection in BDI architectures.
In that work~\citep{leask2018programming}, BDI agents can use system-wide instructions to identify the context in which they operate and this enables them to use a rather reductive notion of reflection called procedural reflection to select deliberation strategies \citep{leask2018programming}.

 Both PRS and BDI architectures can efficiently allow for deliberation, which is distinct from our richer notion of reflection. Kolb's cycle includes the component of Abstract Conceptualisation (see Section \ref{sec:roleofreflection}, Fig. \ref{fig:cycle}), which is missing from the process of procedural reflection. Without this component, deliberation is done without context. Regarding the difference between reflection and deliberation employed in practical reasoning, deliberation is a process for thinking out decisions, whereas reflection is a higher-level process that situates the agent that performs deliberation in a context through Abstract Conceptualisation. Deliberation does not require self-representation through Abstract Conceptualisation, because deliberation can be done at symbol-level, e.g., implementing deliberation strategies and selecting them using a procedural reflection.

\paragraph{Domain Expert Systems} These, such as tutoring expert systems, do not replicate reflection beyond in a rudimentary sense either. For advanced domain expert tutoring systems that are based on architectures like ACT-R \citep{anderson1997act} and that implement some learning theory, they are reflective, but in the procedural sense that we explained above - Lisp style (see Fig.~\ref{fig:Act_R}) \citep{smith1982procedural}. Another issue with systems like ACT-R is that they are architectures for domain expert systems where the environment is part of the system, not agent-based architectures like the critic agent architecture (Fig.~\ref{fig:critic}) where agents act in an observed environment.

To summarise, the ANN architectures discussed above do not allow for reflection to be captured. Conversely, PRS, BDI, and ACT-R do not exclude it; neither do they explicitly describe it.

\section{Building Reflective AI Agents}

In order to make an agent reflective, thus expanding the list of Boden's `sorts of things', we first need an architecture. We must separate out reflection from decision making and action.

Second, we need a suite of reflective cognition processes that may be included depending on the form of reflection desired. A given instance of a reflective agent may have one or more or all of these processes, in line with the differentiated theory approach. We categorise these processes in four tiers:

\textbf{Tier 1 Reflective Agent:} This incorporates models of self and others, and a process to reason using these models in order to ask itself what-if questions concerning generated actions. This enables a Popperian-style consequence engine and reflective governance process, able to evaluate proposed actions in context (acknowledging that context can change) and at least block some actions.

\textbf{Tier 2 Reflective Agent:} Adds processes that learn new reflective models, including incorporating feedback from new experiences into them incrementally. This addition enables Kolb-style reflective experiential learning.

\textbf{Tier 3 Reflective Agent:} Adds a reflective reasoning process that proposes not only a single `optimal' solution, but is ready to present a diversity of possible ways forward -- hypotheses to be tested -- based on different approaches to solving the problem (including safe ways of disengaging from it).

There are many ways an agent may generate proposed actions. These vary in complexity substantially, from simple randomised search (e.g. mutation or exploration) through to heuristic and guided search approaches, up to potentially advanced forms of artificial creativity and strategic planning.

These approaches provide the ability to deliberate about novel possible strategies for action, including in new or potential (imagined) situations, and to evaluate these internally by reasoning with the reflective models.

\textbf{Tier 4 Reflective Agent:} Adds the ability to re-represent existing learnt models in new ways. This facilitates new reasoning possibilities and the potential for new insights. It provides a Gregorian-style ability to change the way the agent reflects.

Third, we need a way of representing the broader context: we need models of prevailing norms, and of the social values associated with the outcomes of different possible actions, and of other higher-level goals that may not be immediately or obviously relevant to the task. 

Note that these components are mostly not new, but it is their novel combination and integration that provides new capability.
Indeed, there are now several decades of work on reflective architectures, including early work like Landauer and Bellman's Wrappings~\citep{Landuer_Bellman:1998}, and Brazier and Treur's~\citep{Brazier_Treur:1995} specification for agents that can reason reflectively about information states. More recently, Blum et al's Consequence Engine architecture~\citep{Blum_et_al:2018}, the EPiCS architecture~\citep{lewis_computer_2015} and the LRA-M architecture~\citep{kounev_notion_2017}, are all aimed explicitly at achieving computational self-awareness through reflection.

On this broader point, self-awareness, often considered as the capacity to be `the object of one's own attention'~\citep{Morin:2006}, has long been targeted as a valuable property for computational systems to possess~\citep{McCarthy:1999,Mitchell:2005}, owing to the value of its functional role and evolutionary advantage in biological organisms~\citep{Lage:2022}.
Computational forms of self-awareness require reflective processes that access, build, and operate on self-knowledge~\citep{Lewis:2011:SASO,kounev_notion_2017}. This self-knowledge is typically described according to five `levels of self-awareness'~\citep{Lewis:2011:SASO,Lewis:2015:Computer,Lewis:2016:Book} rooted in the work of Neisser~\citep{Neisser:1997}, although may consider many other aspects~\citep{Lewis:2017:Ch3}. In some cases these are trivial self-models, for example a smartphone may have an internal parameter that captures whether its charging port contains moisture. Slightly more complex, the device may learn an internal model of its typical charging behaviour, sufficiently to act meaningfully on, and this may adapt as the battery degrades. In more complex examples still, a cyber-physical system may have a model of available resources discovered at run-time~\citep{Bellman:2020:CPS}.

Learning and reasoning with self-knowledge requires a reflective self-modelling process~\citep{Landauer:2016,Bellman:2017:Ch9} of the type described here.
The exact form of such learning and self-modelling will vary depending on requirements and situation, but some examples include self-modelling based on abstraction from run-time data (e.g.,~\citep{Bellman:2017:Ch9}), or simulation of oneself in the environment~\citep{Blum_et_al:2018,Elhabbash:2021}. As Blum et al demonstrate, such simulations may be used as `consequence engines', similarly to how Hesslow \citep{hesslow2002conscious} describes the ability of the human brain to execute processes of internal cognitive simulation.

\begin{figure}[ht]
    \centering
    \includegraphics[width=0.9\linewidth]{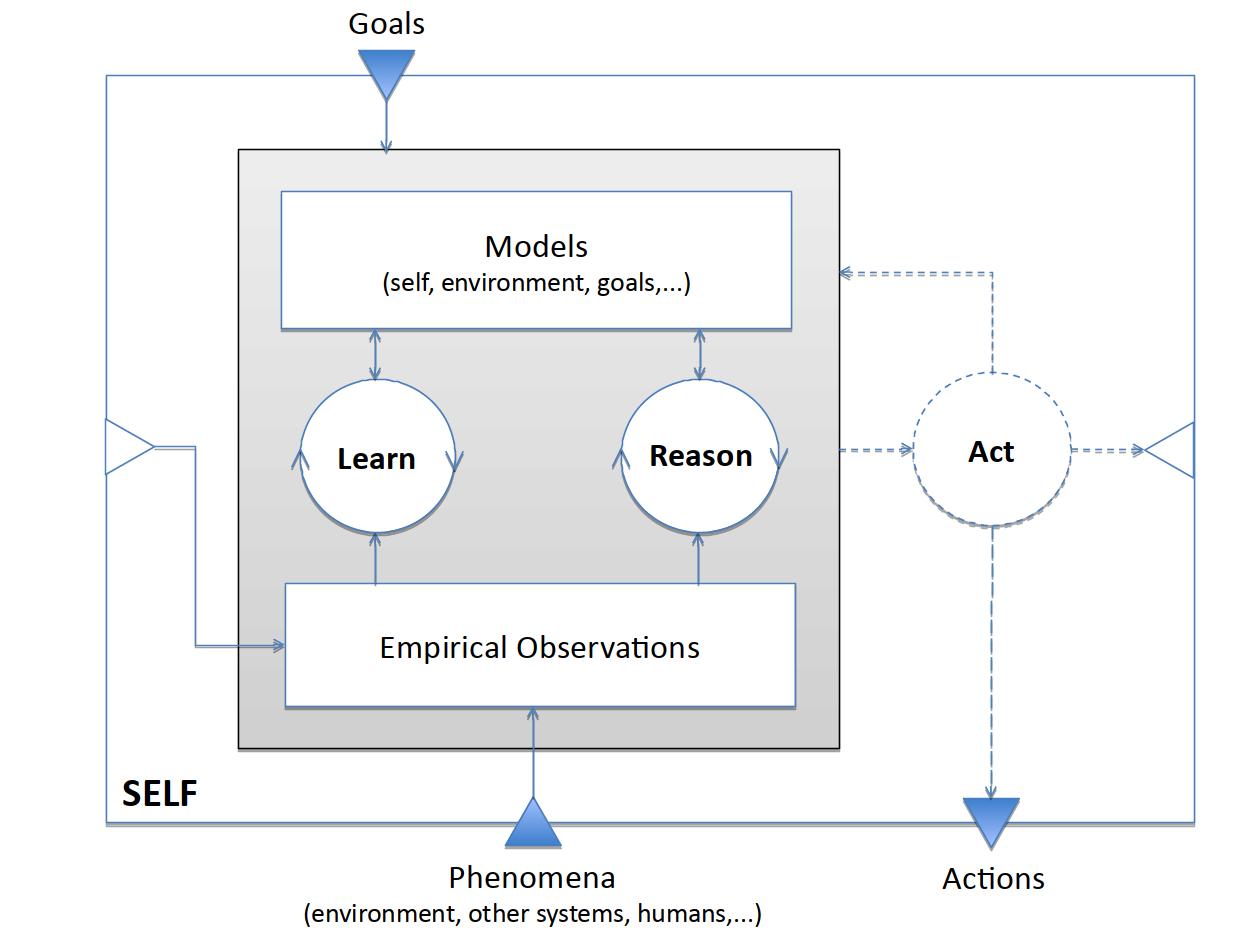}
    \caption{LRA-M Reflective Architecture. Source: \citep{kounev_notion_2017}. The Learn-Reason-Act-Model (LRA-M) model was designed as a reference architecture to capture the essential components of computational self-awareness and their relations. Strictly speaking, acting is considered optional, depicted by the dashed line, though is typically the purpose of the self-awareness in a practical system. The circular arrows signify that learning and reasoning are ongoing processes at run-time, based on streaming data from ongoing observations of the world and oneself. Learning and reasoning also operate on existing internal models, including processes such as re-representation, abstraction, and planning.}
    \label{fig:LRA-M}
\end{figure}
The LRA-M model proposed in ~\cite{kounev_notion_2017} (Figure~\ref{fig:LRA-M}), a commonly used reflective architecture that comes from the area of self-adaptive systems research, captures computational reflection at an abstract level. However, this leaves unclear several aspects associated with agents -- e.g., what process generates the actions? Comparing this with a standard learning-based Critic agent~\citep{russell2021artificial}, we can see the inverse is true: learning and action selection are present, but reflection is not (see Figure~\ref{fig:critic}).

Hence, here we propose one way to integrate the architecture of learning agents with the reflective schema captured by Kounev et al. In this way, a reflective architecture enables information to be abstracted and reasoned with at the meta-level, feeding back to update goals for learning, and to regulate behaviour.

We motivate our choice to base our architecture on Russell \& Norvig's~\citep{russell2021artificial} Critic Agent, and further for using Kounev et al's~\citep{kounev_notion_2017} reflective loop for discussing reflection in AI, since they enjoy broad understanding and acceptance in the domains of agent architecture and computational reflection, respectively. The critic agent, not because it is the best or most state-of-the-art for any particular domain, but because it allows us to illustrate how reflection can be incorporated into a very widely used and understood standard agent architecture. This, we hope, makes the article and argument more accessible. While there are also many reflective loops that we could have chosen, Kounev's is one that enjoys broad support, particularly from the self-adaptive systems community. Indeed, the article that presents that was the result of a large community effort at a Dagstuhl Seminar. Thus, while we acknowledge (and hope) that many other architectures can be paired with other forms of reflective loop, in this article, we use these two as an illustration and first step.

\begin{figure*}[h]
    \centering
    \includegraphics[height=9cm]{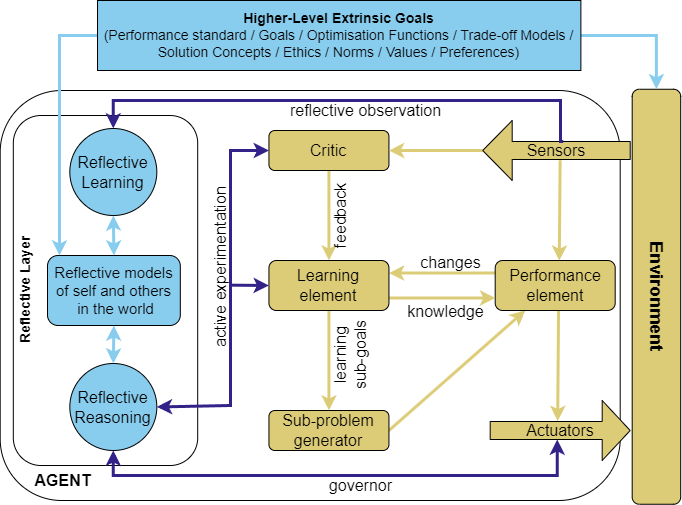}
    \caption{Proposed Reflective Agent Architecture. The yellow elements in the centre and right columns are derived from the Critic Agent architecture~\cite{russell2021artificial}. Elements to the left and above, in blue, are added to form reflective capabilities and are derived from the LRA-M reflective architecture~\cite{kounev_notion_2017}. Connections that integrate these, proposed by us, are depicted in purple and show processes of reflective observation and active experimentation (from~\cite{Kolb:1984}) and behaviour governance (from~\cite{Blum_et_al:2018}).}
    \label{fig:reflective}
\end{figure*}

What we can now see is missing is a simple, generic schema for how reflection relates to existing agent architectures commonly used in modern AI. We propose such an architecture, as a synthesis of Russell and Norvig's \emph{critic agent} and Kounev et al's LRA-M architecture for \emph{reflective computational self-awareness}, illustrated in Figure~\ref{fig:reflective}.


The advantage of an architectural approach is that it describes a separate set of processes, and we know that building systems that self-monitor is easier using an `external layer' style~\citep{weyns2013external}. Further note that what we propose is not an architecture that passes on information from one module to another as is the case in numerous hybrid approaches that aim to marry symbolic and sub-symbolic models \citep{calegari2020integration}. What we propose is a cognitive architecture for reflection which can interpret information before passing it from one module to another. The interpretation of information is dynamic and happens in multiple processes. Below we describe how our proposed architecture ensures information interpretation at different cognitive levels through process loops.

There are many information loops enabled by the addition of this reflective capability. Here, we sketch some of the most obvious and perhaps important ones, particularly those that link to the conceptual discussion above. We categorised the loops according to the corresponding tiers of reflective agents:

\paragraph{Tier 1 Loop -- Governance:}

\begin{itemize}

    \item Loop 1: Governing Behaviour:\\
    Actuators $\rightarrow$ Reflective Reasoning $\rightarrow$ Actuators\\
    \emph{E.g. intervening to prevent an intended action.}\\
    
\end{itemize}

\paragraph{Tier 2 Loops -- Integrating experience and external factors:}
\begin{itemize}

    \item Loop 2: Abstract Conceptualization of Experience\\
    Sensors $\rightarrow$ Reflective Learning $\rightarrow$ Reflective Models $\rightarrow$ Reflective Reasoning $\rightarrow$ Critic.\\
    \emph{E.g. Kolbian experiential learning through conceptualization of new and changing experiences. Also calibrating and correcting exisitng models through new experiences.}\\
    
    \item Loop 3: Learn about and integrate new extrinsic factors into operational goals:\\
    Higher-Level Extrinsic Goals $\rightarrow$ Environment $\rightarrow$ Reflecting Learning $\rightarrow$ Reflective Models $\rightarrow$ Reflective Reasoning $\rightarrow$ Critic.\\
    \emph{E.g. learning about and integrating new external factors, such as social norms, standards, and new user preferences, discovered in the environment, such as signs, verbal instructions, and observation of behaviour.}\\
    
    \item Loop 4: Integrate new design goals into existing reflective models and operational goals:\\
    Higher-Level Extrinsic Goals $\rightarrow$ Reflective models $\rightarrow$ Reflective Reasoning $\rightarrow$ Critic.\\
    \emph{E.g. learning about and integrating new external factors, such as social norms, standards, and new user preferences, discovered in the environment, such as signs and verbal instructions.}
    
\end{itemize}

\paragraph{Tier 3 Loops -- Critique and Imagination:}

\begin{itemize}
 
     \item Loop 5: Active Experimentation to Improve Potential Behaviour\\
    Actuators $\rightarrow$ Reflective Reasoning $\rightarrow$ Critic $\rightarrow$ Learning Element $\rightarrow$ Performance Element $\rightarrow$ Actuators\\
    \emph{E.g. Using the information that an action was intervened upon in order to adapt what the learning element learns, and hopefully avoid the situation in future. Or, creatively proposing novel courses of action and testing hypotheses regarding them.}\\

   \item Loop 6: Reflecting on effectiveness of current operational goals and progress towards them:\\
    Reflective Reasoning $\rightarrow$ Critic $\rightarrow$ Reflective Reasoning.\\
    \emph{E.g. Counterfactual reasoning about current and potential goals, the `what' of operational learning; black-box reasoning about progress towards them, for example asking `am I stuck?' or 'would a different reward function better serve my high-level goals?'}\\

  \item Loop 7: Reflecting on the current mechanisms of learning:\\
    Reflective Reasoning $\rightarrow$ Learning Element $\rightarrow$ Reflective Reasoning.\\
    \emph{E.g. White-box reasoning about current operational learning mechanisms, the `how', for example asking `how am I learning to do this?' and `could I try to learn in a different way?'}\\
 
\end{itemize}

\paragraph{Tier 4 Loop -- Re-Representation:}

\begin{itemize}

 \item Loop 8: Reflective Thinking:\\
    Reflective Reasoning $\rightarrow$ Reflective Models $\rightarrow$ Reflective Reasoning\\
    \emph{E.g. refactoring models, finding and reconciling inconsistencies within and between models, determining areas for further hypothesis testing, re-representing existing conceptual knowledge in new formalisms and abstractions, concept synthesis.}\\

\end{itemize}

One important and powerful insight is that these information loops can be treated as `primitives' and composed to provide additional and more complex cognitive features. For example, the composition of Loops 2 \& 6 could give rise to curiosity-driven behaviour, while adding Loop 8 to this allows the result of the curiosity to be integrated into existing knowledge. Similarly, Loops 1 \& 8 support reflecting on behaviour governance, for example reconciling competing imperatives, assessing the effectiveness of an intervention, or deliberating over an action. Adding Loop 6 to this, permits the deliberation to not only act over potential actions, but over potential directions for future learning.

There are a number of research challenges here. Specifically, there is a need to understand how to operationalize the above information loops, including operational semantics, APIs, and methods for the semantic transformation of information from symbolic to sub-symbolic levels and vice-versa.

\section{How Do We Get There?}

Many of the individual components required to realize Reflective AI already exist. In some cases, the challenge is to integrate these in a purposeful way to achieve the vision set out above. In other cases, there remain important fundamental research challenges. In this Section, we outline some of these in key areas.

\subsection{Reflective Learning} Reflective Learning lies conceptually at the core of the proposed architecture. Fundamentally, learning here provides two forms of modelling capabilities: abstract conceptualisation and simulation, which support reasoning in complementary ways.

Abstract conceptualisation can be described a making sense of observations to form new ideas and theories~\cite{Kolb:1984}. More formally, this includes concept generation in the form of new classification schemes and formal models that represent new theories
and conclusions about events that have been observed. This provides the agent with an interpretation of the experience at a different level of abstraction than the observations themselves. Done computationally, this would enable agents with reasoning processes to make comparisons between competing understandings of a concept, comparing them against future empirical observations and upgrading them through ongoing adaptation. Further, by re-representing these models in new forms, the value of the model to the reasoning process may be similarly upgraded. For example, re-representing a simulation model in closed-form might enable more precise predictions, while re-representing an equation-based model in simulation might extend its predictive scope to capture the outcome of arbitrary or less-well-understood behaviours (e.g.,~\cite{powers2018modelling}). Thus, through abstract conceptualisation and reflective reasoning over these models, the agent has a mechanism for hypothesis generation and testing for the purpose of both future action and cognition.

Note that this process of concept learning and re-representation is distinct from techniques that pass information between cognitive modules operating at the same level of abstraction (cf. \cite{goodfellow2014generative, potter1994cooperative}). If the agent does not have abstraction capabilities, then the interaction more resembles a ping-pong game, where the outcome is each player improving in skill against the other, while the ball and rules retain the same form. An example of this might be a Generative Adversarial Network (GAN)~\citep{goodfellow2014generative} producing ever-better fake faces, but never developing a model to theorise about properties of those faces. However, if these processes can transform the information to new abstractions, then instead a dialectic exists where new understanding can emerge.

However, Abstract Conceptualisation capabilities, have not entered mainstream AI research yet. As illustrated in Figure~\ref{fig:reflective}, such a reflective process could start from Reflective Observation, which takes the data output of the Sensors and passes them to the Reflective Layer, where it uses a Reflective Learning process to transform this data into concepts that can then populate various new or updated self-models. Reasoning over these models can lead to intentional Active Experimentation, targeted at generating new experiences to observe, thus continuing the cycle.

Simulation models support a further form of reasoning, over consequences~\citep{Blum_et_al:2018}. This permits Dennett's `Popperian' mind \citep{dennett2008kinds}, where hypothesis generation and testing can be carried out internally to the cognition of the agent, without requiring the world. For example, in the style of~\citep{hesslow2002conscious}, an agent may build a simulation model in the form of a digital twin of itself in its environment. With sufficient interpretability and accompanied by automated reasoning processes, this may be complemented with an Abstract Conceptualisation, for example, that provides the understanding that the simulation contains an evolutionary stable strategy.

Note that neither the architecture nor the concept of reflective learning prescribe a particular learning algorithm. Many learning techniques can be used. The choice of technique itself is open-ended and can be made to suit the context so long as it adheres to, we posit, two conditions. First, that it is model-based, such that the process of learning produces a model that captures some knowledge about the system and its environment. Interpretable models should be favoured, as the reasoning processes may then operate on these interpretations automatically. Second, that it operates online, such that it can incrementally build and update these models and be used in an anytime fashion.

Indeed, \cite{Lewis:2016:Book} note that online  and lifelong \citep{savage2022learning} learning algorithms are one of the key ingredients in achieving computational self-awareness. They further note that such online learning must be able to deal with concept drift, since both the system and its environment change.
\cite{Wang:2016} show how existing online learning algorithms can be used for reflective self-awareness at different levels, but perhaps most importantly, they intentionally do not propose a preferred online learning paradigm, rather highlighting that empirical results suggest that using different learning techniques according to context can lead to enhanced performance. Complementary examples are presented in a collection edited by Pitt \cite{Pitt:2014:book}, who arrives at a similar conclusion.

In the future, given a mechanism for representing concepts \citep{lieto2021cognitive}, an AI agent could use Kolbian Abstract Conceptualisation to form new concepts and more meaningful models of itself and others in a shared system. Simultaneously, an agent could build simulation models of itself in its environment, to enable Popperian hypothesis testing. Both model forms provide complementary benefits~\citep{powers2018modelling} as forms of reflective modelling for meta-reasoning \citep{brazier1999compositional}, and in different ways, require the ability to learn models on-the-fly~\citep{oltecteanu2019towards}.

\textbf{Research challenge:} There is a need to 
develop mechanisms that learn human- and machine-interpretable conceptual and simulation models from empirical data and semantic information in the world, and further, to develop (unsupervised) methods for this to be done on the fly in a complex environment.

\subsection{Reflective Governance} The proposed architecture captures Socrates's daemon (see Section~\ref{sec:Dangers} above) through a Blum-style governor loop~\citep{Blum_et_al:2018} (also see Section~\ref{sec:Building}), as mediator between Reflective Reasoning and an agent's Actuators. This loop is a process of deliberation at the meta-level. Reflection captures this process and situates it in a context, i.e., in an agent's model of the self and others in the world through Abstract Conceptualisation or simulation. Thus, the system does not need to re-learn its decision model if something in the set of oughts (Higher-level Extrinsic Goals) in its situation changes -- though it might want to, later. It just needs to check the behaviour against them, and occasionally say `no, that's not appropriate; give me an alternative, try a different approach.'

Regarding the ethical nature of this, explicitly ethical agents are nothing new, at least since Moor~\cite{Moor:2009} proposed a way of discerning four different `types'. Indeed, the question of imbuing artificial agents with ethical values was the topic of a special issue of Proceedings of the IEEE. Winfield et al's summary~\citep{Winfield:2019} and Cervantes et al's survey~\citep{cervantes2020artificial} provide an introduction. And indeed, Winfield and colleagues provided an early example~\citep{Blum_et_al:2018} of putting these kinds of `ethical governors' into robots, as consequence engines \citep{winfield2018ethical}; concerns also exist about whether explicit ethical agents are a good idea~\citep{Vanderelst_Winfield:2018}.

\textbf{Research challenge:} There is a need to develop inclusive, participatory methods for capturing values, norms, and preferences in formal, interpretable models that can be translated for use a) in a critic module to drive learning, b) as part of the behaviour governance process, and c) that respects the diversity of interpretation of human values that exists. There is a further need to develop governance and learning processes that adopt these in order to generate and ensure behaviour is aligned with them, as emphasised throughout the Royal Society Special Issue edited by \cite{cath2018governing}.

\subsection{Reflective Deliberation}
Going deeper still, agents could extend the above with reflective deliberation. Reflective agents can deliberate by using Active Experimentation between Reflective Reasoning and Critic (see Figure \ref{fig:reflective}) from time to time to find alternative ways of approaching problems. When considering finding multiple possible diverse and viable courses of action, we can draw on the rich and active research activity on dialogues, practical reasoning and value-based argumentation~\citep{atkinson2005multi,atkinson2007practical,atkinson2016states,atkinson2021value}. These could help us to find new, different solutions, that come at a problem from a novel angle. And when evaluating these alternatives, we may choose to formulate the very notion of what `successful' means according to our values; and in adopting these we must acknowledge that the best action may be a compromise. To instantiate this sort of reflection, agents could employ value-based practical reasoning mechanisms such as action based alternating transition systems with values (AATS+V) or dialogue protocols \citep{sklar2013case}. In turn, these are used to build argument schemes \citep{walton2008argumentation} which agents can use for both reflecting on their possible decisions~\citep{sarkadi2019modelling}, as well as justifying their decisions by providing explanations \citep{mosca2020agent,mosca2021elvira}.

\textbf{Research challenge:} Agents need to be able to perform internal simulations of their actions and check the outcomes of these actions inside their own mind in order to perform deliberation. There is therefore a need to develop semantics and nested abstract models of the world for agent architectures to enable agents to go beyond the procedural reflection of BDI and PRS-like systems, by having the capability to run, analyze, and interpret new simulation models on the fly, according to need.
One idea could be to develop polymorphic simulation models, that can be instantiated into specific simulations based on the learnt concepts and the need.

\subsection{Social Context}
Mentalistic capabilities, as we have explained in the chess example, play an important role in reflecting about one's complex decisions. Again, BDI-like agents can be given both the ability to communicate their decisions to other agents as well as the ability to model the minds of other agents inside their own cognitive architecture in order to better coordinate, or even delegate tasks~\citep{rao1995bdi,sarkadi2018towards}. Social interactions can be modelled and implemented  with dialogue frameworks so that agents can explain and justify their behaviour~\citep{mcburney2007agents,dennis2021explaining}.

Modelling social context is a rich research field. Formal models of norms can be captured using deontic logic; research in normative systems considers the capturing of norms in agents~\citep{criado2011open} and human-robot interactions~\citep{cranefieldnormative}. Social context also includes social values represented in Higher-Level Extrinsic Goals. Solution Concepts~\citep{ficici2004solution} give us one way to formalise these. These can be directly learned at the Reflective Layer by the agent through Reflective Learning. An AI system able to reflect on its actions in terms of social context would need to draw on formal models such as these. Work on normative reasoning in open MAS could play a crucial role, ranging from negotiation between individuals to engineering electronic institutions~\citep{sierra1997framework,sierra2004engineering,pitt2012axiomatization}.

\textbf{Research challenge:} There is a need to develop the semantics and nested abstract models to refine the approaches described in \citep{criado2010bdi,criado2013using,sarkadi2018towards,dennis2021explaining}, by integrating socio-cognitive, communication and normative components inside the instantiated internal simulations. Reflective agents should be able to also simulate the minds and behaviours of other agents and organisations in various contexts where different norms are active, similarly to how Winfield's robots use it to predict the actions of other agents and anticipate the likely consequences of those actions both for themselves and the other agents~\citep{winfield2018experiments}.

\section{Conclusion}

Much research in AI is concerned with breaking a problem down until its constituent parts are solvable; this is important work. Conversely, linking these things together again in an agent-centric fashion to create the sorts of complex mind-like phenomena that motivated us in the first place, is just as crucial.
As we have sketched above, there is a lot to draw on in conceiving and building reflective AI systems. Yet a lot of research remains in understanding how to put together the pieces of the puzzle. Some aspects of reflection are present in the established agent architectures and argumentation models for normative reasoning, deliberation, practical reasoning, and communication. After all, reflection is a crucial component of social interaction, cooperation, and reasoning about what others know and how they might act in different circumstances.

Returning to Weinberg \cite{weinberg1972science}, the idea of Reflective AI is not about providing only scientific answers without any consideration of the broader socio-technical context. Reflective AI will be no silver bullet to the problems raised at the beginning
of this paper, as they are fundamentally trans-scientific in nature.
As such, it presents no excuse to avoid doing AI responsibly, and this would mean falling into the trap of 
what \cite{Oelschaleger} called `the myth of the technological fix'.
Delegating reflective mental capabilities does not nor cannot obviate human responsibility, nor should it distract from it. For example, when building and deploying AI systems, sadly too little attention is still often paid to making them context-sensitive, to understanding stakeholders and operational conditions, to requirements analysis, to understanding bias in data and how it might be amplified, to transparency about training sets, and to interpretability. What we are proposing here is not an either-or.

Instead what we are proposing is a socio-technical mechanism for providing social solutions to social problems, in the context of AI agent technology. To use an analogy, libraries are simply buildings, paper, and databases, that are built by people and enable us to enlighten, inform, and provide pleasure to the population at large. Reflective agents could be a set of methods, tools, and technologies that enable us to contextualise, socialise, put sensitivity into, enrich, and build trust with AI technology.
In doing so, this agenda aims to present a step towards a more complete, less unbalanced conceptualisation of AI systems that allows for more deliberate, careful, and trustworthy technology, if we want to take it.

\bibliographystyle{unsrt}  
\bibliography{references}  

\begin{thebibliography}{10}

\bibitem{Boden:2016:AI}
Margaret~A Boden.
\newblock {\em AI: {Its} Nature and Future}.
\newblock Oxford University Press, 2016.

\bibitem{Mayor:2018}
Adrienne Mayor.
\newblock {\em Gods and Robots: Myths, Machines, and Ancient Dreams of
  Technology}.
\newblock Princeton University Press, 2018.

\bibitem{dignum2020agents}
Virginia Dignum and Frank Dignum.
\newblock {Agents are dead. Long live agents!}
\newblock In {\em Proceedings of the 19th International Conference on
  Autonomous Agents and MultiAgent Systems}, pages 1701--1705, 2020.

\bibitem{sun2001cognitive}
Ron Sun.
\newblock Cognitive science meets multi-agent systems: A prolegomenon.
\newblock {\em Philosophical psychology}, 14(1):5--28, 2001.

\bibitem{lieto2021cognitive}
Antonio Lieto.
\newblock {\em Cognitive design for artificial minds}.
\newblock Routledge, 2021.

\bibitem{Pitt:2014:book}
Jeremy Pitt, editor.
\newblock {\em The Computer After Me}.
\newblock Imperial College Press / World Scientific, 2014.

\bibitem{tine2009uncovering}
M~Tine.
\newblock {\em Uncovering a differentiated Theory of Mind in children with
  autism and Asperger syndrome}.
\newblock PhD thesis, Boston College, USA, 2009.

\bibitem{Bellman_1978}
Richard Bellman.
\newblock {\em {An Introduction to Artificial Intelligence: Can Computers
  Think?}}
\newblock Boyd \& Fraser, San Francisco, 1978.

\bibitem{Monett_et_al:2020}
Dagmar Monett, Colin W.~P. Lewis, Kristinn~R. Thórisson, Joscha Bach, Gianluca
  Baldassarre, Giovanni Granato, Istvan S.~N. Berkeley, François Chollet,
  Matthew Crosby, Henry Shevlin, John Fox, John~E. Laird, Shane Legg, Peter
  Lindes, Tomáš Mikolov, William~J. Rapaport, Raúl Rojas, Marek Rosa, Peter
  Stone, Richard~S. Sutton, Roman~V. Yampolskiy, Pei Wang, Roger Schank, Aaron
  Sloman, and Alan Winfield.
\newblock Special issue ``on defining artificial intelligence'' -- commentaries
  and author's response.
\newblock {\em Journal of Artificial General Intelligence}, 11:1--100, 2020.

\bibitem{Lewis_Marsh:2021}
Peter~R. Lewis and Stephen Marsh.
\newblock What is it like to trust a rock? a functionalist perspective on trust
  and trustworthiness in artificial intelligence.
\newblock {\em Cognitive Systems Research}, 72:33--49, 2021.

\bibitem{Amazon_AI:2018}
Reuters.
\newblock {Amazon ditched AI recruiting tool that favored men for technical
  jobs}.
\newblock {\em The Guardian.}, Oct. 11, 2018, 2018.

\bibitem{Plutarch}
Heinz-Günther Nesselrath, Donald Russell, George Cawkwell, Werner Deuse, John
  Dillon, Heinz-Günther Nesselrath, Robert Parker, Christopher Pelling, and
  Stephan Schröder, editors.
\newblock {\em On the daimonion of Socrates: Plutarch}.
\newblock SAPERE. Mohr Siebeck GmbH and Co. KG, 2010.

\bibitem{Plato_Shorey}
{Plato (translated by Paul Shorey)}.
\newblock {\em Plato in Twelve Volumes}, volume 5 \& 6.
\newblock Harvard University Press, Cambridge, MA, USA, 1969.

\bibitem{dennett2013role}
Daniel~C Dennett.
\newblock {\em The role of language in intelligence}.
\newblock Walter de Gruyter, 2013.

\bibitem{hesslow2002conscious}
Germund Hesslow.
\newblock Conscious thought as simulation of behaviour and perception.
\newblock {\em Trends in cognitive sciences}, 6(6):242--247, 2002.

\bibitem{hesslow2012current}
Germund Hesslow.
\newblock The current status of the simulation theory of cognition.
\newblock {\em Brain research}, 1428:71--79, 2012.

\bibitem{schon}
Donald~A. Schön.
\newblock {\em The Reflective Practitioner: How Professionals Think In Action}.
\newblock Basic Books, 1984.

\bibitem{weinberg1972science}
Alvin~M Weinberg.
\newblock Science and trans-science.
\newblock {\em Science}, 177(4045):211--211, 1972.

\bibitem{Kolb:1984}
David.~A. Kolb.
\newblock {\em Experiential learning: Experience as the source of learning and
  development}.
\newblock Prentice-Hall, Englewood Cliffs, N.J., USA, 1984.

\bibitem{russell2021artificial}
Stuart Russell and Peter Norvig.
\newblock Artificial intelligence: A modern approach, global edition 4th.
\newblock {\em Foundations}, 19:23, 2021.

\bibitem{goodfellow2014generative}
Ian Goodfellow, Jean Pouget-Abadie, Mehdi Mirza, Bing Xu, David Warde-Farley,
  Sherjil Ozair, Aaron Courville, and Yoshua Bengio.
\newblock Generative adversarial networks.
\newblock {\em Communications of the ACM}, 63(11):139--144, 2020.

\bibitem{georgeff1987reactive}
Michael~P Georgeff and Amy~L Lansky.
\newblock Reactive reasoning and planning.
\newblock In {\em AAAI}, volume~87, pages 677--682, 1987.

\bibitem{anderson1997act}
John~R Anderson, Michael Matessa, and Christian Lebiere.
\newblock {ACT-R: A theory of higher level cognition and its relation to visual
  attention}.
\newblock {\em Human--Computer Interaction}, 12(4):439--462, 1997.

\bibitem{mcculloch1943logical}
Warren~S McCulloch and Walter Pitts.
\newblock A logical calculus of the ideas immanent in nervous activity.
\newblock {\em The bulletin of mathematical biophysics}, 5(4):115--133, 1943.

\bibitem{rosenblatt1958perceptron}
Frank Rosenblatt.
\newblock The perceptron: a probabilistic model for information storage and
  organization in the brain.
\newblock {\em Psychological review}, 65(6):386, 1958.

\bibitem{lecun2015deep}
Yann LeCun, Yoshua Bengio, and Geoffrey Hinton.
\newblock Deep learning.
\newblock {\em Nature}, 521(7553):436--444, 2015.

\bibitem{samek2021explaining}
Wojciech Samek, Gr{\'e}goire Montavon, Sebastian Lapuschkin, Christopher~J
  Anders, and Klaus-Robert M{\"u}ller.
\newblock Explaining deep neural networks and beyond: A review of methods and
  applications.
\newblock {\em Proceedings of the IEEE}, 109(3):247--278, 2021.

\bibitem{Sloman_Chrisley:2003}
Aaron Sloman and Ron Chrisley.
\newblock Virtual machines and consciousness.
\newblock {\em Journal of Consciousness Studies}, 10:133--172, 2003.

\bibitem{Sloman:2013}
Aaron Sloman.
\newblock Virtual machine functionalism: The only form of functionalism worth
  taking seriously in philosophy of mind.
\newblock 2013.

\bibitem{Sloman:1996:Rock}
Aaron Sloman.
\newblock What is it like to be a rock?
\newblock 1996.

\bibitem{dennett1975law}
Daniel~C Dennett.
\newblock Why the law of effect will not go away.
\newblock {\em Journal for the Theory of Social Behaviour}, 5:169--187, 1975.

\bibitem{smith1984reflection}
Brian~Cantwell Smith.
\newblock Reflection and semantics in lisp.
\newblock In {\em Proceedings of the 11th ACM SIGACT-SIGPLAN symposium on
  Principles of programming languages}, pages 23--35, 1984.

\bibitem{smith1982procedural}
Brian~Cantwell Smith.
\newblock {\em Procedural reflection in programming languages}.
\newblock PhD thesis, Massachusetts Institute of Technology, 1982.

\bibitem{rao1995bdi}
Anand~S Rao, Michael~P Georgeff, et~al.
\newblock {BDI agents: From theory to practice}.
\newblock In {\em ICMAS}, volume~95, pages 312--319, 1995.

\bibitem{de2020bdi}
Lavindra De~Silva, Felipe Meneguzzi, and Brian Logan.
\newblock Bdi agent architectures: A survey.
\newblock In {\em Proceedings of the 29th International Joint Conference on
  Artificial Intelligence (IJCAI), 2020, Jap{\~a}o.} International Joint
  Conferences on Artificial Intelligence, 2020.

\bibitem{leask2018programming}
Sam Leask and Brian Logan.
\newblock Programming agent deliberation using procedural reflection.
\newblock {\em Fundamenta Informaticae}, 158(1-3):93--120, 2018.

\bibitem{Landuer_Bellman:1998}
Christopher Landauer and Kirstie~L. Bellman.
\newblock Wrappings for software development.
\newblock In {\em Proceedings of the Thirty-First Hawaii International
  Conference on System Sciences}, volume~3, pages 420--429 vol.3, 1998.

\bibitem{Brazier_Treur:1995}
Frances Brazier and Jan Treur.
\newblock Formal specification of reflective agents.
\newblock In {\em IJCAI ‘95 Workshop on Reflection, M. Ibrahim, ed.
  Montreal}, pages 103--112, 1995.

\bibitem{Blum_et_al:2018}
Christian Blum, Alan~FT Winfield, and Verena~V Hafner.
\newblock Simulation-based internal models for safer robots.
\newblock {\em Frontiers in Robotics and AI}, 4:74, 2018.

\bibitem{lewis_computer_2015}
Peter~R. Lewis, Arjun Chandra, Funmilade Faniyi, Kyrre Glette, Tao Chen, Rami
  Bahsoon, J~im~Torresen, and Xin Yao.
\newblock Architectural aspects of self-aware and self-expressive computing
  systems.
\newblock {\em IEEE Computer}, 48:62--70, 2015.

\bibitem{kounev_notion_2017}
Samuel Kounev, Peter Lewis, Kirstie Bellman, Nelly Bencomo, Javier Camara, Ada
  Diaconescu, Lukas Esterle, Kurt Geihs, Holger Giese, Sebastian Göetz, Paola
  Inverardi, Jeffrey Kephart, and Andrea Zisman.
\newblock The notion of self-aware computing.
\newblock In Samuel Kounev, Jeffrey~O. Kephart, Aleksandar Milenkoski, and
  Xiaoyun Zhu, editors, {\em Self-Aware Computing Systems}, pages 3--16.
  Springer, 2017.

\bibitem{Morin:2006}
Alain Morin.
\newblock Levels of consciousness and self-awareness: A comparison and
  integration of various neurocognitive views.
\newblock {\em Consciousness and Cognition}, 15:358--71, 2006.

\bibitem{McCarthy:1999}
John McCarthy.
\newblock Making robots conscious of their mental states.
\newblock In {\em Machine Intelligence 15, Intelligent Agents [St. Catherine's
  College, Oxford, July 1995]}, page 3–17. Oxford University, 1999.

\bibitem{Mitchell:2005}
Melanie Mitchell.
\newblock Self-awareness and control in decentralized systems.
\newblock In {\em Metacognition in Computation}, pages 80--85. AAAI Spring
  Symposium, 2005.

\bibitem{Lage:2022}
Caio~A. Lage, De~Wet Wolmarans, and Daniel~C. Mograbi.
\newblock An evolutionary view of self-awareness.
\newblock {\em Behavioural Processes}, 194:104543, 2022.

\bibitem{Lewis:2011:SASO}
Peter~R. Lewis, Arjun Chandra, Shaun Parsons, Edward Robinson, Kyrre Glette,
  Rami Bahsoon, Jim Torresen, and Xin Yao.
\newblock {A Survey of Self-Awareness and Its Application in Computing
  Systems}.
\newblock In {\em Proceedings of the International Conference on Self-Adaptive
  and Self-Organizing Systems Workshops (SASOW)}, pages 102--107. IEEE Computer
  Society, 2011.

\bibitem{Lewis:2015:Computer}
Peter~R. Lewis, Arjun Chandra, Funmilade Faniyi, Kyrre Glette, Tao Chen, Rami
  Bahsoon, Jim Torresen, and Xin Yao.
\newblock Architectural aspects of self-aware and self-expressive computing
  systems.
\newblock {\em IEEE Computer}, 48:62--70, 2015.

\bibitem{Lewis:2016:Book}
Peter~R. Lewis, Marco Platzner, Bernhard Rinner, Jim T{\o}rresen, and Xin Yao,
  editors.
\newblock {\em Self-Aware Computing Systems: An Engineering Approach}.
\newblock Springer, 2016.

\bibitem{Neisser:1997}
Ulric Neisser.
\newblock The roots of self-knowledge: Perceiving self, it, and thou.
\newblock {\em Annals of the New York Academy of Science}, 818:19--33, 1997.

\bibitem{Lewis:2017:Ch3}
Peter~R. Lewis, Kirstie~L. Bellman, Christopher Landauer, Lukas Esterle, Kyrre
  Glette, Ada Diaconescu, and Holger Giese.
\newblock Towards a framework for the levels and aspects of self-aware
  computing systems.
\newblock In Samuel Kounev, Jeffrey~O. Kephart, Aleksandar Milenkoski, and
  Xiaoyun Zhu, editors, {\em Self-Aware Computing Systems}, pages 3--16.
  Springer, 2017.

\bibitem{Bellman:2020:CPS}
K.~Bellman, C.~Landauer, N.~Dutt, L.~Esterle, A.~Herkersdorf, A.~Jantsch,
  N.~TaheriNejad, P.~R. Lewis, M.~Platzner, and K.~Tammem\"{a}e.
\newblock Self-aware cyber-physical systems.
\newblock {\em ACM Transactions on Cyber-Physical Systems}, 4(4), 2020.

\bibitem{Landauer:2016}
Christopher Landauer and Kirstie~L. Bellman.
\newblock Reflective systems need models at run time.
\newblock In Sebastian G{\"{o}}tz, Nelly Bencomo, Kirstie~L. Bellman, and
  Gordon~S. Blair, editors, {\em Proceedings of the 11th International Workshop
  on Models@run.time co-located with 19th International Conference on Model
  Driven Engineering Languages and Systems {(MODELS} 2016), Saint Malo, France,
  October 4, 2016}, volume 1742 of {\em {CEUR} Workshop Proceedings}, pages
  52--59. CEUR-WS.org, 2016.

\bibitem{Bellman:2017:Ch9}
Kirstie~L. Bellman, Christopher Landauer, Phyllis Nelson, Nelly Bencomo,
  Sebastian Götz, Peter~R. Lewis, and Lukas Esterle.
\newblock Self-modeling and self-awareness.
\newblock In Samuel Kounev, Jeffrey~O. Kephart, Aleksandar Milenkoski, and
  Xiaoyun Zhu, editors, {\em Self-Aware Computing Systems}, pages 3--16.
  Springer, 2017.

\bibitem{Elhabbash:2021}
Abdessalam Elhabbash, Rami Bahsoon, Peter Tino, Peter~R Lewis, and Yehia
  Elkhatib.
\newblock Attaining meta-self-awareness through assessment of
  quality-of-knowledge.
\newblock In {\em 2021 IEEE International Conference on Web Services (ICWS)},
  pages 712--723. IEEE Computer Society, 2021.

\bibitem{weyns2013external}
Danny Weyns, M~Usman Iftikhar, and Joakim S{\"o}derlund.
\newblock Do external feedback loops improve the design of self-adaptive
  systems? a controlled experiment.
\newblock In {\em 2013 8th International Symposium on Software Engineering for
  Adaptive and Self-Managing Systems (SEAMS)}, pages 3--12. IEEE, 2013.

\bibitem{calegari2020integration}
Roberta Calegari, Giovanni Ciatto, and Andrea Omicini.
\newblock {On the integration of symbolic and sub-symbolic techniques for XAI:
  A survey}.
\newblock {\em Intelligenza Artificiale}, 14(1):7--32, 2020.

\bibitem{powers2018modelling}
Simon~T Powers, Anik{\'o} Ek{\'a}rt, and Peter~R Lewis.
\newblock Modelling enduring institutions: The complementarity of evolutionary
  and agent-based approaches.
\newblock {\em Cognitive Systems Research}, 52:67--81, 2018.

\bibitem{potter1994cooperative}
Mitchell~A Potter and Kenneth~A De~Jong.
\newblock A cooperative coevolutionary approach to function optimization.
\newblock In {\em International Conference on Parallel Problem Solving from
  Nature}, pages 249--257. Springer, 1994.

\bibitem{dennett2008kinds}
Daniel~C Dennett.
\newblock {\em Kinds of minds: Toward an understanding of consciousness}.
\newblock Basic Books, 2008.

\bibitem{savage2022learning}
Neil Savage.
\newblock Learning over a lifetime.
\newblock {\em Nature}, 2022.

\bibitem{Wang:2016}
Shuo Wang, Georg Nebehay, Lukas Esterle, Kristian Nymoen, , and Leandro~L.
  Minku.
\newblock Common techniques for self-awareness and self-expression.
\newblock In Peter~R. Lewis, Marco Platzner, Bernhard Rinner, Jim T{\o}rresen,
  and Xin Yao, editors, {\em Self-Aware Computing Systems: An Engineering
  Approach}, pages 113--142. Springer, 2016.

\bibitem{brazier1999compositional}
Frances~MT Brazier and Jan Treur.
\newblock Compositional modelling of reflective agents.
\newblock {\em International Journal of Human-Computer Studies},
  50(5):407--431, 1999.

\bibitem{oltecteanu2019towards}
Ana-Maria Olte{\c{t}}eanu, Mikkel Sch{\"o}ttner, and Arpit Bahety.
\newblock Towards a multi-level exploration of human and computational
  re-representation in unified cognitive frameworks.
\newblock {\em Frontiers in psychology}, 10:940, 2019.

\bibitem{Moor:2009}
James~H. Moor.
\newblock Four kinds of ethical robots.
\newblock {\em Philosophy Now}, 72:12--14, 2009.

\bibitem{Winfield:2019}
Alan~F. Winfield, Katina Michael, Jeremy Pitt, and Vanessa Evers.
\newblock Machine ethics: The design and governance of ethical ai and
  autonomous systems [scanning the issue].
\newblock {\em Proceedings of the IEEE}, 107(3):509--517, 2019.

\bibitem{cervantes2020artificial}
Jos{\'e}-Antonio Cervantes, Sonia L{\'o}pez, Luis-Felipe Rodr{\'\i}guez,
  Salvador Cervantes, Francisco Cervantes, and F{\'e}lix Ramos.
\newblock Artificial moral agents: A survey of the current status.
\newblock {\em Science and Engineering Ethics}, 26(2):501--532, 2020.

\bibitem{winfield2018ethical}
Alan~FT Winfield and Marina Jirotka.
\newblock Ethical governance is essential to building trust in robotics and
  artificial intelligence systems.
\newblock {\em Philosophical Transactions of the Royal Society A: Mathematical,
  Physical and Engineering Sciences}, 376(2133):20180085, 2018.

\bibitem{Vanderelst_Winfield:2018}
Dieter Vanderelst and Alan F.~T. Winfield.
\newblock The dark side of ethical robots.
\newblock In {\em AAAI/ACM Conference on AI Ethics and Society}, pages
  317--322, 2018.

\bibitem{cath2018governing}
Corinne Cath.
\newblock Governing artificial intelligence: ethical, legal and technical
  opportunities and challenges.
\newblock {\em Philosophical Transactions of the Royal Society A: Mathematical,
  Physical and Engineering Sciences}, 376(2133):20180080, 2018.

\bibitem{atkinson2005multi}
Katie Atkinson, Trevor Bench-Capon, and Peter McBurney.
\newblock {Multi-Agent Argumentation for eDemocracy}.
\newblock In {\em EUMAS}, pages 35--46, 2005.

\bibitem{atkinson2007practical}
Katie Atkinson and Trevor Bench-Capon.
\newblock Practical reasoning as presumptive argumentation using action based
  alternating transition systems.
\newblock {\em Artificial Intelligence}, 171(10-15):855--874, 2007.

\bibitem{atkinson2016states}
Katie Atkinson and Trevor Bench-Capon.
\newblock States, goals and values: Revisiting practical reasoning.
\newblock {\em Argument \& Computation}, 7(2-3):135--154, 2016.

\bibitem{atkinson2021value}
Katie Atkinson and Trevor Bench-Capon.
\newblock Value-based argumentation.
\newblock {\em Journal of Applied Logics}, 8(6):1543--1588, 2021.

\bibitem{sklar2013case}
Elizabeth~I Sklar, Mohammad~Q Azhar, Simon Parsons, and Todd Flyr.
\newblock A case for argumentation to enable human-robot collaboration.
\newblock {\em Proceedings of Autonomous Agents and Multiagent Systems (AAMAS),
  St Paul, MN, USA}, 2013.

\bibitem{walton2008argumentation}
Douglas Walton, Christopher Reed, and Fabrizio Macagno.
\newblock {\em Argumentation schemes}.
\newblock Cambridge University Press, 2008.

\bibitem{sarkadi2019modelling}
{\c{S}}tefan Sarkadi, Alison~R Panisson, Rafael~H Bordini, Peter McBurney,
  Simon Parsons, and Martin Chapman.
\newblock Modelling deception using theory of mind in multi-agent systems.
\newblock {\em AI Communications}, 32(4):287--302, 2019.

\bibitem{mosca2020agent}
Francesca Mosca, {\c{S}}tefan Sarkadi, Jose~M Such, and Peter McBurney.
\newblock Agent {EXPRI}: Licence to explain.
\newblock In {\em International Workshop on Explainable, Transparent Autonomous
  Agents and Multi-Agent Systems}, pages 21--38. Springer, 2020.

\bibitem{mosca2021elvira}
Francesca Mosca and Jose Such.
\newblock Elvira: an explainable agent for value and utility-driven multiuser
  privacy.
\newblock In {\em International Conference on Autonomous Agents and Multiagent
  Systems (AAMAS)}, 2021.

\bibitem{sarkadi2018towards}
{\c{S}}tefan Sarkadi, Alison~R Panisson, Rafael~H Bordini, Peter McBurney, and
  Simon Parsons.
\newblock Towards an approach for modelling uncertain theory of mind in
  multi-agent systems.
\newblock In {\em International Conference on Agreement Technologies}, pages
  3--17. Springer, 2018.

\bibitem{mcburney2007agents}
Peter McBurney and Michael Luck.
\newblock The agents are all busy doing stuff!
\newblock {\em IEEE Intelligent Systems}, 22(4):6--7, 2007.

\bibitem{dennis2021explaining}
Louise~A Dennis and Nir Oren.
\newblock Explaining bdi agent behaviour through dialogue.
\newblock In {\em Proc. of the 20th International Conference on Autonomous
  Agents and Multiagent Systems (AAMAS 2021)}. International Foundation for
  Autonomous Agents and Multiagent Systems (IFAAMAS), 2021.

\bibitem{criado2011open}
Natalia Criado, Estefania Argente, and V~Botti.
\newblock Open issues for normative multi-agent systems.
\newblock {\em AI communications}, 24(3):233--264, 2011.

\bibitem{cranefieldnormative}
Stephen Cranefield and Bastin Tony~Roy Savarimuthu.
\newblock Normative multi-agent systems and human-robot interaction.
\newblock In {\em Workshop on Robot Behavior Adaptation to Human Social Norms
  (TSAR)}, pages 1--3, 2021.

\bibitem{ficici2004solution}
Sevan~Gregory Ficici.
\newblock {\em Solution Concepts in Coevolutionary Algorithms}.
\newblock PhD thesis, Brandeis University, 2004.

\bibitem{sierra1997framework}
Carles Sierra, Nick~R Jennings, Pablo Noriega, and Simon Parsons.
\newblock A framework for argumentation-based negotiation.
\newblock In {\em International Workshop on Agent Theories, Architectures, and
  Languages}, pages 177--192. Springer, 1997.

\bibitem{sierra2004engineering}
Carles Sierra, Juan~Antonio Rodriguez-Aguilar, Pablo Noriega, Marc Esteva, and
  Josep~Lluis Arcos.
\newblock Engineering multi-agent systems as electronic institutions.
\newblock {\em European Journal for the Informatics Professional}, 4(4):33--39,
  2004.

\bibitem{pitt2012axiomatization}
Jeremy Pitt, Julia Schaumeier, and Alexander Artikis.
\newblock Axiomatization of socio-economic principles for self-organizing
  institutions: Concepts, experiments and challenges.
\newblock {\em ACM Transactions on Autonomous and Adaptive Systems (TAAS)},
  7(4):1--39, 2012.

\bibitem{criado2010bdi}
Natalia Criado, Estefania Argente, and V~Botti.
\newblock A bdi architecture for normative decision making.
\newblock In {\em Proceedings of the 9th International Conference on Autonomous
  Agents and Multiagent Systems: volume 1-Volume 1}, pages 1383--1384, 2010.

\bibitem{criado2013using}
Natalia Criado.
\newblock Using norms to control open multi-agent systems.
\newblock {\em AI Communications}, 26(3):317--318, 2013.

\bibitem{winfield2018experiments}
Alan~FT Winfield.
\newblock Experiments in artificial theory of mind: From safety to
  story-telling.
\newblock {\em Frontiers in Robotics and AI}, 5:75, 2018.

\bibitem{Oelschaleger}
Max Oelschlaeger.
\newblock The myth of the technological fix.
\newblock {\em The Southwestern Journal of Philosophy}, 10(1):43--53, 1979.

\end{thebibliography}


\end{document}